\documentclass{article}
\usepackage{spconf,amsmath,graphicx}

\usepackage{enumitem}
\setlist{nosep, leftmargin=14pt}

\usepackage{mwe} 

\title{Learning the Domain Specific Inverse NUFFT for Accelerated Spiral MRI using Diffusion Models}
\name{Trevor J. Chan\textsuperscript{1}, Chamith S. Rajapakse\textsuperscript{2}}
\address{1. Department of Bioengineering, University of Pennsylvania, Philadelphia, United States \\ 2. Department of Radiology, University of Pennsylvania, Philadelphia, United States}
\begin{document}
%
\maketitle
\begin{abstract}
Deep learning methods for accelerated MRI achieve state-of-the-art results but largely ignore additional speedups possible with noncartesian sampling trajectories. To address this gap, we created a generative diffusion model-based reconstruction algorithm for multi-coil highly undersampled spiral MRI. This model uses conditioning during training as well as frequency-based guidance to ensure consistency between images and measurements. Evaluated on retrospective data, we show high quality (structural similarity \textgreater 0.87) in reconstructed images with ultrafast scan times (0.02 seconds for a 2D image). We use this algorithm to identify a set of optimal variable-density spiral trajectories and show large improvements in image quality compared to conventional reconstruction using the non-uniform fast Fourier transform. By combining efficient spiral sampling trajectories, multicoil imaging, and deep learning reconstruction, these methods could enable the extremely high acceleration factors needed for real-time 3D imaging. 
\end{abstract}


\section{Introduction}
\label{sec:intro}

Despite the numerous advantages of MRI, inherent physical and hardware constraints cap acquisition speed and lead to long scan times. This creates myriad downstream obstacles, including low patient compliance, inefficient resource allocation, and image motion artifacts, among others. For this reason, methods to accelerate MRI acquisition have been and continue to be an active area of research. Acceleration can be achieved through a combination of using more efficient scanning sequences and reducing the number of measurements made in the frequency space of the image, k-space. 

Considering the former, successful techniques for faster scanning include radial and spiral imaging methods, which exploit the unequal distribution of information across k-space by sampling lower frequencies more densely \cite{blum1987fast,winkelmann2006optimal}.

Considering the latter, acquiring fewer measurments in k-space rewards a decrease in scan time proportional to the ratio of undersampling, but comes at the cost of image quality. Sampling below the Nyquist limit introduces ambiguities during reconstruction which manifest as artifacts in the final image. Reconstruction algorithms must therefore leverage additional information, including multicoil data in the case of parallel imaging, and sparse priors, in the case of compressed sensing, in order to resolve these ambiguities \cite{sodickson1997simultaneous,lustig2007sparse}.

A third more recent approach to undersampled MRI reconstruction lies in deep learning methods, which essentially learn a set of image priors and use these to regularize solutions to the ill-posed reconstruction problem \cite{oscanoa2023deep}. Within this category, diffusion models stand out for producing state-of-the art results on image reconstruction tasks for faster scanning, motion correction, noise reduction, and others \cite{song2021solving, johnson2020improving, cui2023spirit, aali2023solving}. Despite this, the vast majority of deep learning approaches, and to our knowledge, all diffusion-based approaches to image reconstruction, focus on Cartesian-sampled MRI, missing out on potential acceleration gains attained by using more efficient non-Cartesian sampling trajectories. \textbf{This work addresses this gap by introducing a diffusion model-based method for trajectory-agnostic undersampled reconstruction of multicoil spiral MRI.} 
\linebreak \linebreak
\textit{\textbf{Contributions:}}
\begin{itemize}
  \item \textit{Creation of a novel multi-conditioning strategy for solving the inverse non-uniform fast Fourier transform (nufft) using a learned conditional score function and weak frequency guidance during sampling}
  \item \textit{Efficient hyperparameter search of the joint trajectory-reconstruction space and identification of optimal sampling trajectories}
  \item \textit{Retrospective acquisition and reconstruction of a 2D, 256x256 pixel, 22x22 cm\textsuperscript{2} image with a readout duration of 0.02 seconds.}
\end{itemize}

\begin{figure*}[ht]
\begin{minipage}[b]{1.0\linewidth}
  \centering
  \centerline{\includegraphics[width=17 cm]{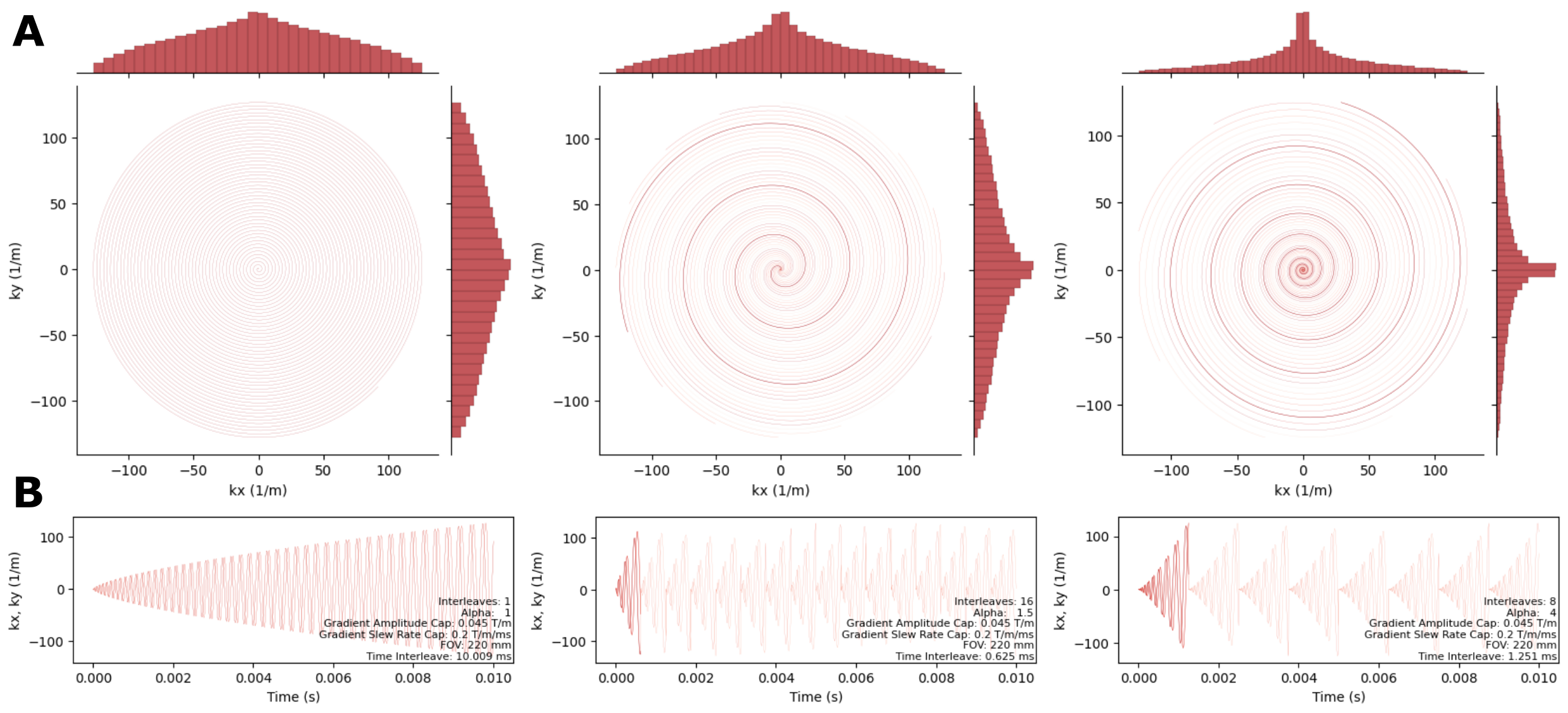}}
 \vspace{-0.3cm}
\end{minipage}
\caption{Example trajectories (A) and the corresponding readout gradients in $k_x$ and $k_y$ (B). All trajectories shown cover the frequency space of a 256x256 image and have a readout duration of 10.0 ms.}
\vspace{-0.3cm}
\label{fig:trajectories}
\end{figure*}

\section{Background}
\label{sec:background}

Canonical score-based models consider the mapping between a known distribution of independently and identically distributed samples of gaussian noise and an observed, but unknown distribution of data \(p(x)\). These distributions bookend a Langevin diffusion process described by the following stochastic differential equation representing the trajectory of a sample from our data distribution into a sample from our noise distribution:
\begin{equation}
dx = f(x,t) dt + g(t) d\mathrm{w}.
\end{equation}
Here, functions $f(\cdot,t)$ and $g(\cdot)$ are the drift and diffusion coefficiets of $x(t)$ respectively, and \textbf{w} is a standard Wiener process, or Brownian motion.

In order to generate a novel sample from our data distribution, we can generate a random noise vector and attempt to solve the reverse-time SDE, but this is generally intractable. However, we can approximate this process by estimating the noise-conditioned score function, $\nabla_x \log p_t(x)$, which computes the likelihood of a sample \(x\) existing between the noise and image distributions. With this, the reverse-time SDE becomes:
\begin{equation}
dx = \left[ f(x,t) - g(t)^2 \nabla_x \log p
_t(x) \right]dt+ g(t)d\bar{\mathrm{w}}
\end{equation}
The score function can be trained using a score matching with Langevin dynamics algorithm \cite{song2020score, karras2022elucidating}.

\section{Methods}
\label{sec:methods}

This research study was conducted retrospectively using human subject data made available in open access by \cite{knoll2020fastmri}. Ethical approval was not required.

\subsection{Data}
We use the NYU FastMRI dataset \cite{knoll2020fastmri} consisting of 6970 fully sampled 2D brain scans on hardware ranging from 4 to 24 coils. For training and testing, we consider axial T2 weighted turbo spin echo sequences characterized by the following sequence parameters: scan time=140 s, TR=6 s, TE=113 ms, slices=30, slice thickness=5 mm, field of view=22 cm, matrix size=320x320. Effective scan time for a 2D slice at 256\textsuperscript{2} resolution is $140 \mathrm{s} / 320 * 256 / 30 \approx 3.7 \mathrm{s}$. As this data is initially acquired using Cartesian sequences, we simulated spiral acquisition by retrospectively interpolating in k-space to attain complex-valued measurements along generated spiral trajectories.

\subsection{Generating spiral trajectories}
Following Kim et al. \cite{kim2003simple}, we consider spiral trajectories of the form
\begin{equation}
    k(\tau) =  \int_{0}^{\tau}  \frac{1}{\rho(\phi)} d\phi e^{j\omega \tau}\;\;\approx\;\; \lambda \tau^\alpha e^{j \omega \tau}.
\end{equation}
Here, $\rho$ denotes sampling density, $\tau$ is a function of time, $\phi$ is angular position, $\omega = 2\pi n$ is frequency, with $n$ the number of turns in k-space, $\lambda$ a scaling factor equal to $matrix\;size / (2*FOV)$, and $\alpha$ is a bias term for oversampling the center of k-space relative to the edges. Solving this parametric equation under the constraints of capped gradient slew rate and capped gradient amplitude yields gradients ($g_x(t)$ and $g_y(t)$) as well as a spiral trajectory in the $k_x$,$k_y$-plane (figure 1). In doing so, we can tune sampling parameters to control for factors such as read out duration and dwell time, while varying the number of interleaves and ratio of low-to-high frequency oversampling.

\begin{figure*}[ht]
\begin{minipage}[b]{1.0\linewidth}
  \centering
  \centerline{\includegraphics[width=12 cm]{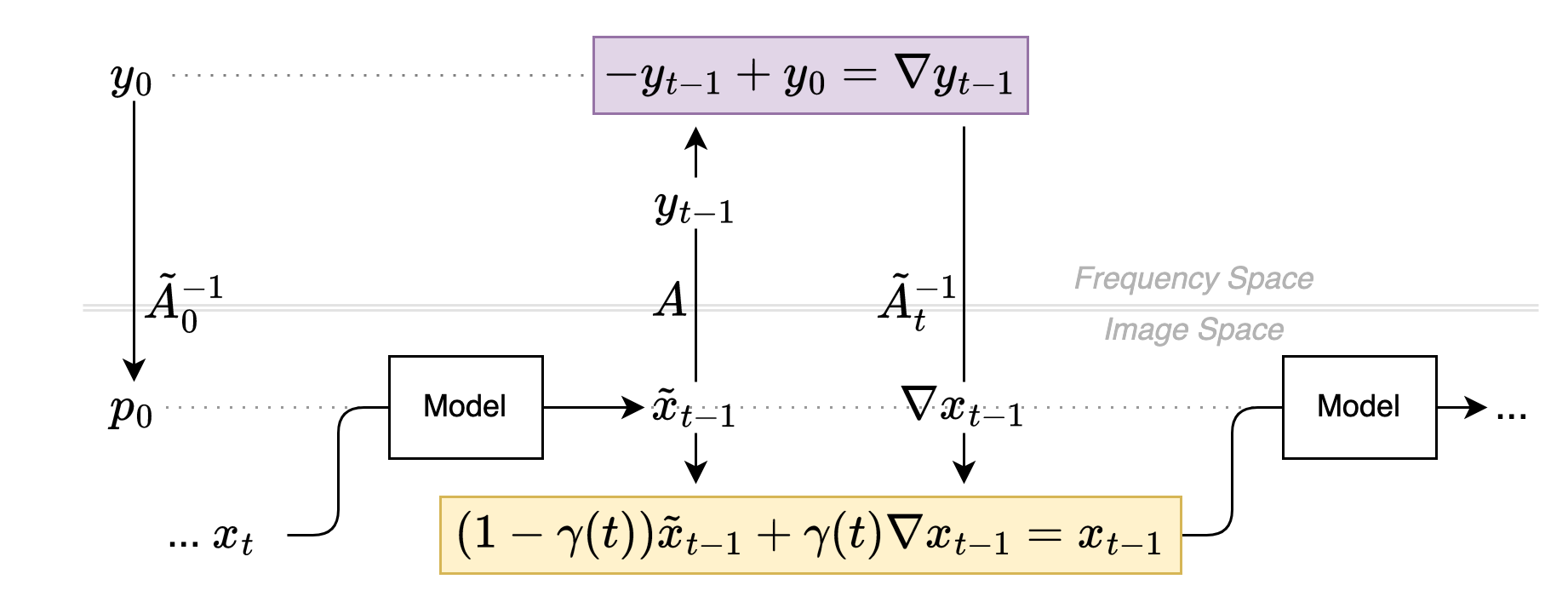}}
 \vspace{-0.3cm}
\end{minipage}
\caption{Given measurements $y_0$, reconstruction follows a modified diffusion sampling process. At each timestep, a noisy latent $x_t$ is concatenated with a prior $p_0$ and passed to the denoising model to obtain $\tilde x_{t-1}$. To enforce consistency with $y_0$, we compute a frequency gradient $\nabla y_{t-1}$ and solve for the image gradient using a modified iterative inverse nufft (section \ref{equation:modified_nufft}). A weighted sum of $x_{t-1}$ and $\nabla x_{t-1}$ yields the corrected image $x_{t-1}$. This is repeated until $t=0$.}
\label{fig:sampling}
\end{figure*}

\begin{figure*}[!h]
\begin{minipage}[b]{1.0\linewidth}
  \centering
  \centerline{\includegraphics[width=17.5cm]{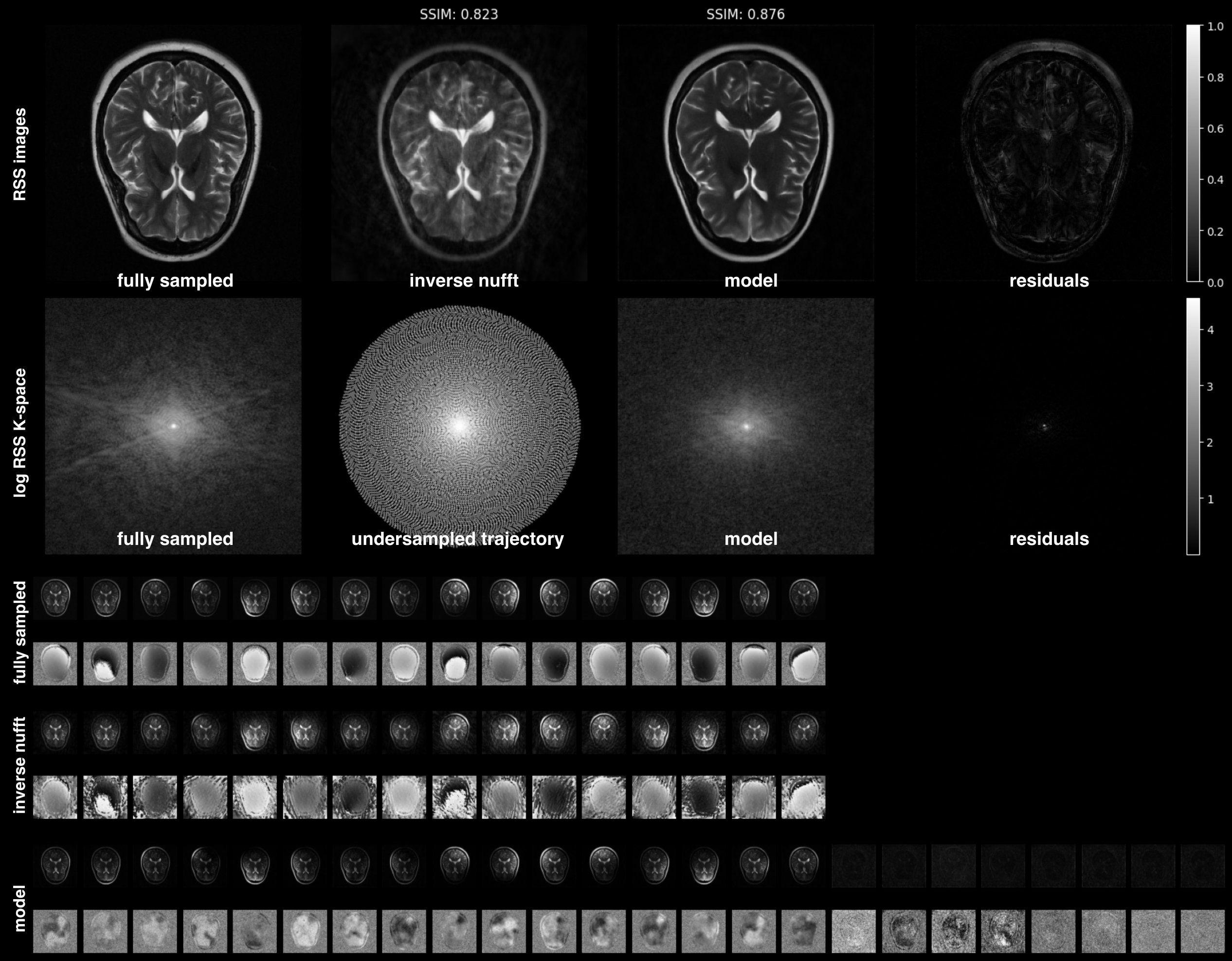}}
 \vspace{-0.3cm}
\end{minipage}
\caption{Representative reconstruction results for a single 2D 16 coil image. Retrospective k-space data was sampled with an optimized 23 interleave sequence with a total readout duration of 0.02 s. Rows 1 and 2 show the RSS-reconstructed images and log-scaled k-space magnitudes for the ground truth, inverse nufft, and proposed model reconstructions. Below are the individual coil magnitude and phase images for the fully sampled image, the inverse nufft reconstructions, and the model predictions.}
\vspace{-0.3cm}
\label{fig:results}
\end{figure*}

\begin{figure}[!b]
\begin{minipage}[b]{1.0\linewidth}
  \centering
  \centerline{\includegraphics[width=8.55cm]{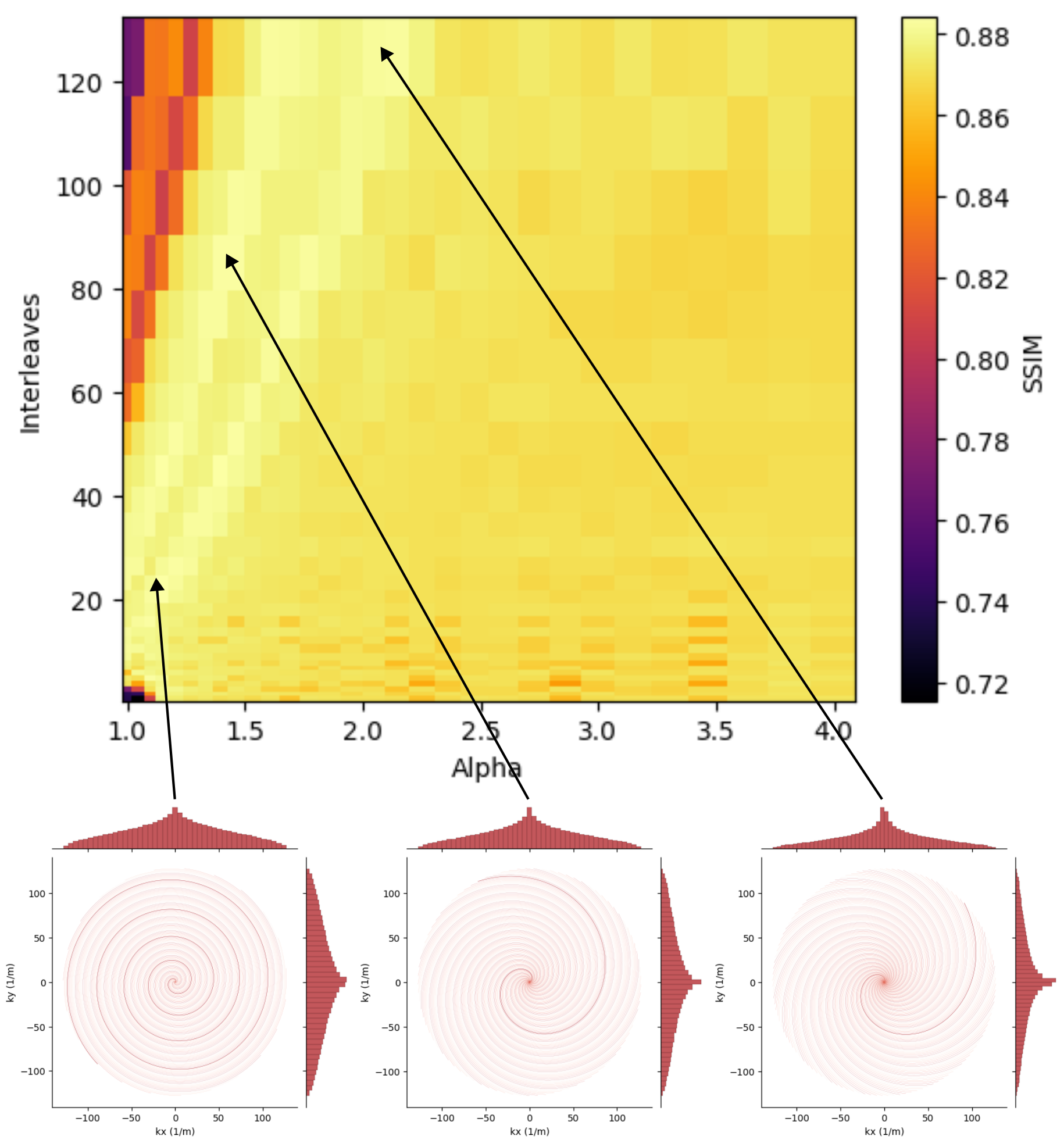}}
 \vspace{-0.3cm}
\end{minipage}
\caption{We performed a grid hyperparameter search over a 2D trajectory space. We fixed readout duration at 0.02 seconds and varied the number of interleaves from 1 to 125 and alpha from 1 to 4. Based on structural similarity of the model-reconstructed images, we found multiple trajectories that yield improved image quality. In comparison, the naive Archimedean spiral, corresponding to 1 interleave and $\alpha=1$, performs very poorly.}
\vspace{-0.3cm}
\label{fig:gridsearch}
\end{figure}

\begin{figure*}[t]
\begin{minipage}[b]{1.0\linewidth}
  \centering
  \centerline{\includegraphics[width=17.8cm]{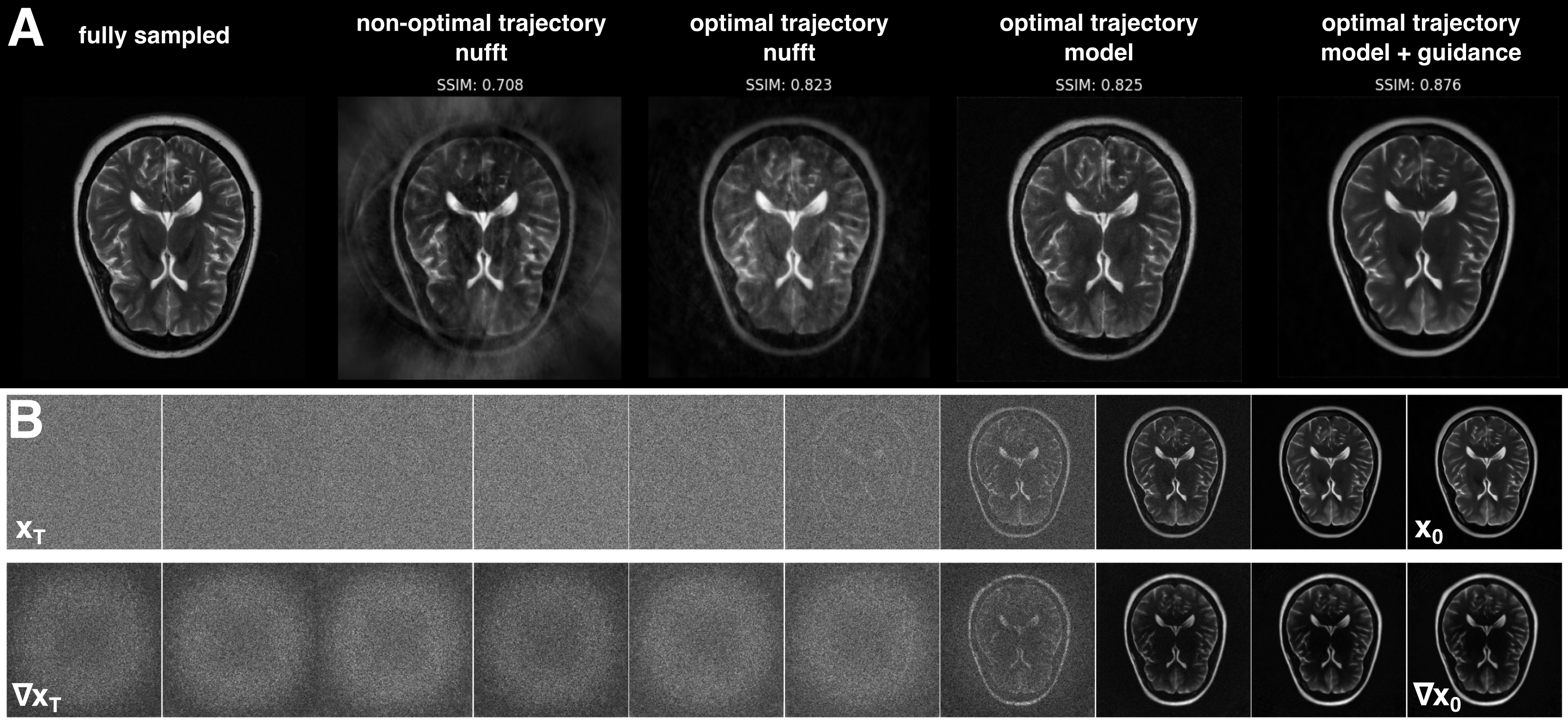}}
 \vspace{-0.3cm}
\end{minipage}
\caption{(A) Effect of sampling trajectory optimization, model reconstruction without frequency guidance, and model reconstruction with frequency guidance. For the non-optimized trajectory, we used a single interleave Archimedean spiral with a readout duration of 0.02 s. The optimized trajectory uses a 23 interleave, $\alpha=1.23$ sequence with an identical readout duration. (B) Snapshots of the image latent $x_t$ and the gradient signal $\nabla x_t$ taken during a diffusion sampling process.}
\vspace{-0.3cm}
\label{fig:ablations}
\end{figure*}

\subsection{Image reconstruction is inverse problem solving}
MRI undersampled acquisition amounts to measuring an unknown signal $x$ through some imperfect sampling function $A$: $y = Ax + \epsilon$. Here, $y$ is the measured multicoil k-space data, and $A$ is the non-uniform fourier transform. $\epsilon$ is measurement noise and exists in the same domain as the $y$; in MRI, noise is gaussian-distributed across the real and imaginary components of $y$ for each coil.

Reconstruction is an ill-posed inverse problem of recovering an image signal $x$ from a set of incomplete k-space measurements $y$. As $x$ and $y$ exist in different domains, $x$ is hidden behind a sampling operator $A$. Solving this problem necessitates prior knowledge. In our case, we learn an underlying conditional distribution of images and seek to reconstruct samples from this distribution consistent with the measurements. Information is supplied in two forms: first, we learn a conditional score function $\nabla_x \log p_t(x_t|\tilde{A}^{-1} y_0)$, where $y_0$ is the measurement in frequency space and $\tilde{A}^{-1}$ is an approximate inverse of $A$, in our case the inverse nufft solved iteratively using conjugate gradients. We find that adding this supervision during training helps to constrain the model when faced with a large number of input image channels and the periodic ambiguity inherent when operating on complex numbers.

Second, we use frequency space gradients to weakly guide the sampling process. At each time step during sampling, we compute the forward nufft of an uncorrected $\tilde{x}_{t-1}$ and take a difference between that and the measurements $y_0$. To minimize this difference, the approximate gradient in image space is calculated by solving a modified approximate inverse nufft $\tilde{A_t}^{-1}$, which corrects for low frequency biases and applies time step-dependent noising determined by the noise schedule $\sigma(t)$, which is necessary for sampling with langevin diffusion. 
\begin{equation}
    \tilde{A_t}^{-1}x_t = \frac{\tilde{A}^{-1}x_t}{c_1e^{-c_2r^2}} + \mathcal{N}(0, \sigma(t)^2)
\end{equation}
\label{equation:modified_nufft}
Following \cite{karras2022elucidating}, we choose a linear noise schedule and observe that the underlying ordinary differential equation describing transit from latent to image is locally linear, so summation of $\tilde{x}_{t-1}$ and $\nabla x_{t-1}$ to obtain a frequency-corrected image $x_{t-1}$ is akin to gradient descent (figure 2). 

In practice, due to the non-invertibility of the nufft, imperfections in the approximate inverse nufft bleed into the final image reconstruction, introducing artifacts and reducing quality. To avoid this, we anneal the guidance signal following an empirically chosen linear schedule $\gamma(t)=\beta(1-t)$, ensuring strong guidance at the outset of sampling and minimal artifacts at the end of sampling. A consequence of this choice is that we do not strongly enforce that $Ax_0 = y_0$ at time 0.

\section{Results}
\label{sec:results}

Model reconstruction performance was evaluated on a held-out test dataset. Test trajectories have a fixed readout duration of 0.02 seconds, in which time the measurements needed to reconstruct a 256x256 pixel, 22x22 cm\textsuperscript{2} 2D image are acquired. Reconstructed image quality was scored using structural similarity (SSIM) (figure 3).

To investigate the effect choosing different scanning trajectories has on the quality of reconstructed images, we also perform a grid hyperparameter search of spiral trajectories with a fixed readout duration of 0.02 seconds (figure 4) and varying $\alpha$ and interleaves. Surprisingly, the common 'naive' trajectory, a single interleave Archimedean spiral, corresponding to $\alpha=1$, performs very poorly when sampled below the Nyquist limit. Trajectories which perform better tended to lie along two logarithmic curves roughly characterized by $\alpha=1.33\log(0.39 \, interleaves)$ and $\alpha=0.87\log(0.54 \, interleaves)$.

Finally, we conduct an ablation study to disentangle the effects of optimal sampling trajectory without model reconstruction, model reconstruction without frequency guidance, and model reconstruction with frequency guidance (figure 5). We find that all three contribute to noticeable increases in image quality, both visual, and quantitative based on SSIM. The combination of choosing an optimal trajectory, performing model reconstruction with conditioning, and using annealed frequency guidance results in large improvements in image quality, up to and exceeding a 0.15 boost in SSIM.

\section{Discussion}
\label{sec:discussion}

While initial results are promising, the main limitation of this project is the reliance on retrospective, Cartesian-sampled data. Implementing the sequences outlined in this work will likely require customizing spiral sequences so as to match the contrast and signal of the original Cartesian sequences, which will constrain the space of realizable trajectories. Until a dataset of raw non-Cartesian MRI data becomes available, this will continue to be an obstacle. For a similar reason, it is difficult to make head-to-head comparisons between the original sequence and the proposed sequences without prospective validation. For this reason, the immediate task is to acquire prospective data with custom sequences and use it to validate image reconstruction. Currently, a concrete direct comparison is between the proposed sequences and their Nyquist-sampled counterparts, which run roughly 3x longer.

Apart from the short-term task of matching sequence parameters between spiral and Cartesian sequences, the choice of spiral sequence leaves a considerable amount of flexibility even within the space of optimized interleave and density pairs identified above. Variation in number of interleaves, and by extension the duration of a single interleave, allows for tailoring of sequence contrast, signal, and speed to task-specific requirements. An area in which these sequences could provide additional benefit, even outside of sheer acceleration, would be in imaging tissues with a very short T2\textsuperscript{*}, as acceleration within an interleave allows for proportionally more data acquisition to occur before signal has decayed.

\section{Conclusion}
\label{ssec:conclusion}

Here we introduce a new method and show preliminary results for reconstructing spiral MRI using a diffusion model. Combining multicoil imaging, spiral scanning, and undersampling enables dramatically faster imaging speeds. Applications of this work are widespread; in addition to the numerous typical benefits associated with faster scanning, including better patient compliance and fewer motion artifacts, these methods have the potential to reach the extremely high acceleration factors necessary to achieve high resolution real-time 3D imaging.

\section{Acknowledgments}
\label{sec:acknowledgments}
No funding was received for conducting this study. The authors have no relevant financial or non-financial interests to disclose.

\bibliographystyle{IEEEbib}
\bibliography{refs}

\end{document}